\documentclass[runningheads]{llncs}
\usepackage{graphicx}
\usepackage{caption}
\usepackage{array}
\usepackage{booktabs}
\usepackage{tabularx}
\usepackage{makecell}
\usepackage{pbox}
\usepackage{xcolor}
\usepackage{adjustbox}
\usepackage{color, colortbl}
\definecolor{Gray}{gray}{0.9}

\usepackage{pgfplots}
\pgfplotsset{compat=newest}
\usepackage{tikz}
\usetikzlibrary{decorations.pathmorphing}

\usepackage{amsmath}
\usepackage{pifont}
\newcommand{\cmark}{\ding{51}}%
\newcommand{\xmark}{\ding{55}}%

\usepackage[utf8]{inputenc}
\usepackage{pgfplots}
\usepgfplotslibrary{groupplots,dateplot}
\usetikzlibrary{patterns,shapes.arrows}
\pgfplotsset{compat=newest}

\begin{document}
\title{Estimating Presentation Competence using Multimodal Nonverbal Behavioral Cues}
\titlerunning{Estimating Presentation Competence}
%
\author{Ömer Sümer\inst{1} \and
Cigdem Beyan\inst{4} \and
Fabian Ruth\inst{2} \and
Olaf Kramer\inst{2} \and
Ulrich Trautwein\inst{3} \and 
Enkelejda Kasneci\inst{1} 
}
\authorrunning{Sümer and Beyan et al.}
%
\institute{Department of Computer Science, University of Tübingen \and
Department of General Rhetoric, University of Tübingen \and
Hector Research Institute of Education Sciences and Psychology \\ University of Tübingen, Tübingen, Germany
\and
Department of Information Engineering and Computer Science, \\ University of Trento, Povo (Trento), Italy \\
\email{\{name.surname\}@uni-tuebingen.de,\{name.surname\}@unitn.it}}
\maketitle              
\begin{abstract}
Public speaking and presentation competence plays an essential role in many areas of social interaction in our educational, professional, and everyday life. Since our intention during a speech can differ from what is actually understood by the audience, the ability to appropriately convey our message requires a complex set of skills. Presentation competence is cultivated in the early school years and continuously developed over time. One approach that can promote efficient development of presentation competence is the automated analysis of human behavior during a speech based on visual and audio features and machine learning. Furthermore, this analysis can be used to suggest improvements and the development of skills related to presentation competence. In this work, we investigate the contribution of different nonverbal behavioral cues, namely, facial, body pose-based, and audio-related features, to estimate presentation competence. The analyses were performed on videos of 251 students while the automated assessment is based on manual ratings according to the Tübingen Instrument for Presentation Competence (TIP). Our classification results reached the best performance with early fusion in the same dataset evaluation (accuracy of 71.25\%) and late fusion of speech, face, and body pose features in the cross dataset evaluation (accuracy of 78.11\%). Similarly, regression results performed the best with fusion strategies. 
\end{abstract}
\section{Introduction}\label{introduction} Public speaking requires a high caliber of eloquence and persuasion in order to convey the speaker's objective while also captivating their audience. Above all, public speaking is essential to many educational and professional aspects of life, e.g., a successful thesis defense, teaching a lecture, securing a job offer, or even presenting your research at a conference. Moreover, in the context of digital transformation and with increasing online presence (e.g., online teaching courses), the demand for tutorials related to the development of presentation competence is expanding rapidly.  For example, the non-profit educational organization Toastmasters International\footnote[1]{https://www.toastmasters.org/}, which teaches public speaking through a worldwide network of clubs, currently has more than 358K members.

Besides the actual content of a speech (the verbal cues), multiple nonverbal cues, such as prosody, facial expressions, hand gestures, and eye contact, play a significant role in engaging with, convincing, and influencing the audience \cite{Riggio:1986,Knapp:2013}. Various public speaking performance rubrics \cite{Carlson:1995,Thomson:2002,Morreale:2007,Schreiber:2012} have been used by teachers and professors to manually asses the competence of a speech. Although the rubrics above consider a speaker's nonverbal behavior, some do not differentiate between types of nonverbal behavior (acoustic or visual). For instance, Schreiber et al. \cite{Schreiber:2012} include nonverbal cues as a single item: ''demonstrating nonverbal behavior that reinforces the message''. While it is certainly possible for a human annotator to utilize high-inference questions when rating a performance, by employing machine learning we can further investigate fine-grained nonverbal behaviors individually and provide speakers with detailed feedback to improve their presentation skills.

With this motivation in mind, our work employs a recently proposed assessment rubric, the Tübingen Instrument for Presentation competence (TIP), whose items represent nonverbal cues in detail. Having different items for behavioral cues, such as posture, gesture, facial expressions, eye contact, and audio traits, allows for a better explainability of the strengths and weaknesses of a public speech. In contrast to the sole assessment of a speech in previous works, we can, in this way, infer the underlying behavioral factors, and enable an automated assessment, which can become an asset in (self) training.

Besides their time-consuming nature, manual assessments are prone to subjectivity. Although a proper training and simultaneous rating by multiple raters might help overcome this limitation, relying on human raters limits the number of assessments that can be done at a certain time. To tackle these problems, automatic public speaking competence estimation is necessary. Some studies in the social computing domain have therefore investigated automated assessment with regard to audio-based nonverbal features (NFs) \cite{Rao2017,Pfister:2011,Luzardo:2014}, video-based NFs \cite{Sharma:2018,Chen:2014}, or with a multimodal approach as in \cite{Wortwein:2015:a,Wortwein:2015:b,Haider:2016,Chen:2014,Chen:2015,Ramanarayanan:2015}. Related works that performed automated public speaking competence analysis indicate that there are different types of speeches such as scientific presentations \cite{Sharma:2018,Ramanarayanan:2015,Herbein2018,Curtis2015}, political speeches \cite{Scherer:2012,Cullen:2018}, and video interviews \cite{Rao2017}.

In this study, we compare three major sources of nonverbal communication: \textit{i)} speech, \textit{ii)} face (including head pose and gaze), and \textit{iii)} body pose, as well as the fusion of these sources, to assess public speaking competence.
The experimental analyses were 
conducted on informational, scientific presentations performed using visual aids and in front of a two-person audience.\footnote[2]{Different terms, such as public speaking or presentation, were used to refer a person speaking in front of a group. In our study, we prefer using presentation and presentation competence, however, to retain the original terminology used in the previous works.} 

Our main contributions are as follows:
\begin{itemize}
    \item We conduct an in-depth analysis of nonverbal features extracted from the face, body pose, and speech for automatic presentation competency estimation in videos when features per modality are used alone or when they are fused. The features' effectiveness is examined when they are extracted from the whole video (so-called global features) and extracted from shorter video segments (so-called local features) for classification and regression tasks. These analyses are performed for a person-independent within the same dataset, and a person-specific cross-dataset setting.
    
    \item Previous studies in the computational domain used different and non-structured evaluation instruments for presentation competence. This study validates a recently proposed evaluation metric, Tübingen Instrument for Presentation Competence (TIP). We also present Youth Presents Presentation Competence Dataset and conduct the first analysis to compare various nonverbal features and learning models in this data using TIP measures.
    
    \item 3-minute scientific presentations are emerging as an academic genre \cite{Hu:2018,Rowley-Jolivet:2019}. Such short scientific presentations are publicly available on the internet and can also be used in combination with automated methods to estimate presentation competence. We initially validated the usability of short scientific presentations for this purpose. 
\end{itemize}
The remainder of this paper is organized as follows. Section \ref{literature_review} reviews related work on automated public speaking competence estimation and assessment rubrics. Section \ref{dataset} describes the data sets and presentation competence instrument used in our analysis. In Section \ref{section:approach}, we describe the proposed method in detail. Experimental analyses, the results of classification, regression and correlation analyses and cross-data experiments are provided in Section \ref{section:experimental_analysis}. Lastly, we conclude the paper and discuss the limitations and future work in Section \ref{section:conclusion}.

\section{Literature Review}\label{literature_review}
Investigating the relationship between acoustic/visual nonverbal features (NFs) and public speaking performance can contribute to the development of an automated platform for speaker training and/or assessment. Below, we review social computing literature for public speaking performance analysis. There are several studies, but they are restricted to a single type of NFs, lack the adequate sample sizes, or have no differentiation in terms of speech types. Additionally, different assessment rubrics used in psychology and education domains to measure presentation quality are discussed.

\subsection{Estimating Presentation Competence}
Early on, Rosenberg and Hirschberg \cite{Rosenberg:2005} found correlations between acoustic and lexical features of charismatic speech. Their defined acoustic features were the mean, standard deviation, and maximum of the fundamental frequency (f0) and speaking rate. Lexical features were defined as the number of first-person pronouns, etc.  
Later, Strangert and Gustafson \cite{Strangert:2008} found that speakers with more dynamic f0 range were perceived more positively during political debates. Although these works~\cite{Rosenberg:2005,Strangert:2008} provide preliminary research into public speaking competence, they are limited by subjective rubrics, small datasets, and few features. In addition to acoustic features (e.g., prosody and voice quality), Scherer et al. \cite{Scherer:2012} examined body, head, and hand motion-based NFs to investigate their influence on the perception of political speeches. From eye-tracking data, they found that human observers mainly concentrate on speakers' faces when viewing audio-visual recordings, but concentrate on speakers' bodies and gestures when viewing visual-only recordings. 

In the education domain, the \emph{Multimodal Learning Analytics} (MLA) data corpus comprises of 40 oral presentations of students from the challenge workshop \cite{Ochoa:2014:MLA}, including audiovisual recordings and slides. However, the manual assessment criteria/rubrics used were not published. Using this corpus, Chen et al. \cite{Chen:2014} applied a Support Vector Machine (SVM) and gradient boosting to the combination of audio intensity, pitch, the displacement of body parts detected by Kinect sensors, head pose, and slide features (e.g., the number of pictures or grammatical errors).
Using the same data, Luzardo et al. \cite{Luzardo:2014} utilized the slide features (e.g., the number and size of text, pictures, tables) together with the audio features (e.g., pause fillers, pitch average, pitch variation) and applied an instance-based classifier. However, their approach is not suitable for public speeches without visual aids and neglects speakers' nonverbal features. Although these studies used manually extracted verbal features, they promoted efforts for semi-automatic speaking performance assessment. 

Moving towards automated presentation assessment, Haider et al. \cite{Haider:2016} focused on prosodic and gestural features to categorize presentation quality as poor vs. good. In total, 6376 audio features and 42 statistical features representing hand motions were adapted for the classification of presentation classification. More important, they demonstrated that multimodal NFs perform better than using NFs of each modality alone. Specifically, it was found that presentation quality factors highly correlate with each other. In other words, it is possible to detect visual NFs with prosody features.

Continuing in the direction of multimodal features for automated assessment, Wörtwein et al. \cite{Wortwein:2015:a} developed a model to assess and improve speaker performance. Nine items measuring behavioral indicators (e.g., body pose, the flow of speech, eye contact) were defined, and audiovisual data annotated via crowd-sourcing was proposed. A relative annotation was performed by comparing two videos displayed at the same time. Correlations between extracted NFs and behavioral indicators were shown. The extracted audio-visual NFs were also used to train and make inference with ensemble classifiers.
Conversely, Pfister et al.~\cite{Pfister:2011} claimed that highly persuasive speech requires a display of emotions consistent with verbal content. They applied affective states recognized by audio-based NFs for public speaking skill analysis and achieved
89\% and 61\% classification accuracy on average and within leave-one-speaker-out cross validation, respectively.

To the best of our knowledge, Chen et al.~\cite{Chen:2015}  and Ramanaraynanan~\cite{Ramanarayanan:2015} are the only studies in the literature that utilize the public speaking competence rubric (RSCP) [cite], a well established assessment rubric.
The public speaking performance ratings are automatically estimated using Support Vector Regression (SVR), Random Forest (RF), and generalized linear models. They use the time-aggregated statistics and histogram of co-occurrences of NFs; head pose, gaze, facial expressions, and body locations. The main drawbacks of these studies \cite{Chen:2015,Ramanarayanan:2015} are the evaluation on a limited size of data and poor performance for some items in the rubric.

\begin{table}[ht!] 
	\centering
	\caption{Comparison of Assessment Rubrics for Presentation Competence.}
	\label{table:presentation_instruments}
	\resizebox{\linewidth}{!}{%
	\begin{tabular}{lllllll} 
		\toprule
		Assessment Rubric & \makecell[l]{Target\\ level}   & Item number & \makecell[l]{Seperate \\ items \\ per NFs} & \makecell[l]{Sample \\ (\#speech)}  & \makecell[l]{(Interrater) \\ Reliability} \\
		\midrule
		\makecell[l]{\textbf{Classroom Public} \\ \textbf{Speaking Assessment} \\ Carlson et al. \cite{Carlson:1995}}  & \makecell[l]{higher \\ education} & \makecell[l]{(Form B) \\ 5 items/ \\ 5-point scale}   & \xmark  & 2 &  \makecell[l]{-- Cronbach coefficient: \\ from $.69$ to $.91$ }\\
		\midrule
		\makecell[l]{\textbf{Public Speaking} \\ \textbf{Competency Instrument} \\ Thomson et al. \cite{Thomson:2002} }& \makecell[l]{higher \\ education} & \makecell[l]{20 items/ \\ 5-point scale}  & \xmark  & 1 & n.a. \\
		\midrule
		\makecell[l]{\textbf{Competent Speaker} \\ \textbf{Speech Evaluation Form} \\  Morreale et al. \cite{Morreale:2007} }& \makecell[l]{higher \\ education} & \makecell[l]{8 items/ \\ 3-point scale}   & \xmark  & 12 & \makecell[l]{-- Ebel's coefficient: \\ from $.90$ to $.94$ \\ -- Cronbach coefficient:  \\ from $.76$ to $.84$ } & \\
		\midrule
		\makecell[l]{\textbf{Public Speaking}\\ \textbf{Competence Rubric} \\ Schreiber et al. \cite{Schreiber:2012}} & \makecell[l]{higher \\ education} & \makecell[l]{11 items/ \\ 5-point scale}   & \xmark  & 45-50 & \makecell[l]{ICC: \\ $.54 \leq r \leq.93$}  \\
		\midrule
		\makecell[l]{\textbf{Tübingen Instrument}\\ \textbf{for Presentation Competence} \\ Ruth et al. \cite{Ruth:2019}} & \makecell[l]{high \\ school} & \makecell[l]{22 items/ \\ 4-point scale}   & \cmark & \makecell[l]{161 \emph{(T1)} \\  94 \emph{(T2)}} &  \makecell[l]{-- Cronbach coefficient:  \\ from $.67$ to $.93$ \\ -- ICC $>.60$ for 10 \\ out of 15 items}  \\
		\bottomrule
	\end{tabular}
	}
\end{table}

\subsection{Assessment Rubrics for Presentation Competence}
The ability of an automated system to decipher and report public speaking competence is incredibly valuable. One way to realize this characteristic is to use a systematic rubric that can address each possible NF as separate items. The judgments made using such a rubric can also provide better training data and can help human observers improve their confidence and rate of decision-making \cite{Thomson:2002}.

Carlson and Smith-Howell \cite{Carlson:1995} developed three evaluation forms for informative speeches. They tested these forms on two award-winning presenters' speeches with one speech made intentionally less informative by changing the delivery and content of the speech. These speeches were evaluated by 58 individuals using the evaluation forms. Two of the three forms showed higher inter-reliability (Cronbach's $\alpha=.83$ and $.91$). However, any of these forms include separate items representing NFs individually. Instead, visual NFs are into one item as presentation and delivery of all visual nonverbal cues.

A more recent instrument, namely, the Competent Speaker Speech Evaluation Form \cite{Morreale:2007},
can be used to evaluate speeches in a class environment. It can instruct students about how to prepare and present public speeches, and can generate assessment data for the accountability-related objectives of academic institutions. 
In this form, the acoustic NFs are defined as vocal variety in rate, pitch, and intensity, but are still represented in a single item. Visual NFs are not even defined. This kind of assessment may be suitable for classroom evaluation purposes and training automated algorithms, but it does not help to identify what is ``insufficient'' and can be improved in students' individual presentations.

One of the most comprehensive assessment tools for reporting indicators of objectivity, reliability, and validity is \cite{Schreiber:2012}. This rubric has 11-items (nine core and two optional) with a 5-point scale (\emph{4-advanced, 3-proficient, 2-basic, 1-minimal,} and \emph{0-deficient}). The audio-based and video-based NFs are individually considered as: ``Representing how effective the speaker uses vocal expression and paralanguage\footnote[3]{Paralanguage is the field of study that deals with the nonverbal qualities of speech (i.e. pitch, amplitude, rate, and voice quality).} to engage the audience,'' and ``demonstrating the competence of posture, gestures, facial expressions and eye contact that supports the verbal message,'' respectively. These items are more informative, but NFs have still not been represented individually.

Unlike the aforementioned rubrics, Thomson and Rucker \cite{Thomson:2002} described individual items regarding a speaker’s speech volume, gestures, and eye contact as being relaxed and comfortable as well as voice and body expressiveness. However, this rubric lacks facial expressions and posture features.

In summary, even though these rubrics provide a suitable foundation for public speaking performance assessment, there is an absence of more fine-grained items that represent various NFs separately. A more detailed comparison of the rubrics is presented in Table~\ref{table:presentation_instruments}. In the current study, we use a more detailed rubric, especially for assessing NFs, which is introduced in the next section.

\begin{table}[ht!]
	\caption{Description of Tübingen Instrument for Presentation Competence (TIP) Items.}
	\centering
	\begin{tabular}{p{0.02\columnwidth}p{0.6\columnwidth}} 
		\toprule
		\textbf{Item} & \ \ \ \textbf{Description} \\ 
		\toprule
		& \emph{Addressing the audience} \\ \hline
		1 & ... addresses the audience.\\ 
		2 & ... has a motivating introduction.\\
		3 & ... takes the listeners’ questions and \\
		&     \ \ \  expectations into account. \\ \hline
		& \emph{Structure} \\ \hline
		4 & ... introduces the presentation convincingly.\\
		5 & ... structures transitions convincingly.\\
		6 & ... ends the presentation convincingly  \\
		&  \ \ \  with a conclusion.\\ \hline
		& \emph{Language use} \\ \hline
		7 & ... uses examples to create a tangible portrayal \\
		&  \ \ \   of the topic.\\
		8 & ... uses appropriate sentence structures for oral\\
		&   \ \ \  communication.\\
		9 & ... uses technical terms appropriately.\\ \hline
		& \emph{Body language \& voice} \\ \hline
		10 & ... has an effective posture.\\
		11 & ... employs gestures convincingly.\\
		12 & ... makes eye contact with the audience\\
		& \ \ \ convincingly.\\
		13 & ... uses facial expressions convincingly.\\
		14 & ... uses their voice effectively \\
		& \ \ \ (melody, tempo, volume).\\ 
		15 & ... uses their voice convincingly\\
		&  \ \ \ (articulation, fluency, pauses).\\ \hline
		& \emph{Visual aids} \\ \hline
		16 & ... uses an appropriate amount of visual information.\\
		17 & ... structures visual elements appropriately.\\
		18 & ... constructs an effective interplay \\
		   &  \ \ \ between the speech and visual aids.\\
		19 & ... creates visual aids which are visual attractive. \\
		20 & ... formulated an appropriately clear scientific question. \\ 
		21 & ... appears confident in handling information.\\
		22 & ...'s reasoning is comprehensible.\\
		\bottomrule
	\end{tabular}
	\label{Table:TIP}
\end{table}

\section{Assessment Rubric and Data Sets}\label{dataset}
\subsection{Tübingen Instrument for Presentation Competence}\label{section3.1.TIP}
The items of the Tübingen Instrument for Presentation Competence (TIP) depend on rhetorical theory and cover six faces of presentation competence: \emph{addressing the audience, structure, language use, body language \& voice, visual aids}, and \emph{content credibility}. In total there are 22 TIP items as shown in Table~\ref{Table:TIP}. All items are in a 4-point Likert-type scale (1 = not true to 4 = very true). 

As we aim to investigate the nonverbal behaviors for presentation competence, in the experimental analysis provided in Section \ref{section:approach} we only used the data corresponding to items 10-15 (i.e., body language and voice). How the corresponding ratings are used for regression and classification tasks are described in Section \ref{section:ClassReg}.

\subsection{Youth Presents Presentation Competence Dataset}
The Youth Presents Presentation Competence Dataset was collected during the second (T1) and third-round (T2) of the Youth Presents contest\footnote[3]{\url{https://www.jugend-praesentiert.de/ueber-jugend-praesentiert}}, a nationwide German presentation contest for secondary school students aged 12 to 20. Informed consent was obtained from all students and their parents before the study began, and the study protocol was approved by the ethics committee of the University. Students who submitted their video presentations were first pre-assessed by a jury and then selected for the second round. In this round, they were asked to give a presentation in front of a jury on a scientific topic of their choice. Their presentations were video-recorded and constituted the first set of the Youth Presents (T1). After assessing these presentations, the best performing students were invited some weeks later to the third round. The third round included an exercise presentation under standardized conditions that had no consequences for the contest. These video-recorded presentations constitute the second set of the Youth Presents (T2).

Both sets of the Youth Presents include three-minute presentations in front of a jury consisting of two people. The presenters were using analog visual aids (e.g. poster, object, experiments, notation on the blackboard). In some aspects, the presentation tasks differed between T1 and T2. Relatively speaking, students at T1 had more time to prepare: E.g., they were allowed to make analog visual aids at home and chose the scientific content of their presentation. Students at T2 were assigned the content of their presentation (microplastics in the environment) and had 40 minutes of preparation time. Additionally, they were provided a set of text materials on the topic and visualization materials (i.e., three colored pens and six white papers for a bulletin board). 

Overall, 160 students delivered a presentation in the T1 condition. 91 of those presented a second time at T2. The overall number was 251 videos and the mean age of the students is 15.63 years (std = 1.91).
Each video was rated by four trained raters who were first introduced to the theoretical foundations of presentation competence, familiarized with the rating items, and performed exemplary ratings of video-recorded presentations that were not part of T1 and T2. 
During the training process, the raters discussed their ratings based on anchor examples in order to establish a common understanding of the rating items. The overall training procedure took 36 hours. After the training, each rater assessed all videos independently. The order of the videos was randomized to avoid order effects.

For each TIP item, the interrater reliability was calculated using a two-way, mixed, absolute, average-measures intraclass correlation coefficient (ICC) \cite{McGraw1996}. The results showed that among 22 items given in Table~\ref{Table:TIP}, 15 items at T1 (except items 4, 8, 9, 10, 15, 17, 22) and 14 items in T2 (except 5, 7, 8, 9, 10, 17, 20, 22) exhibited ICCs above $0.60$. High ICC value ($>0.60$) indicates high interrater reliability and implies that the criteria rated similarly across raters.

\begin{figure*}
\begin{center}
\includegraphics[width=0.8\linewidth]{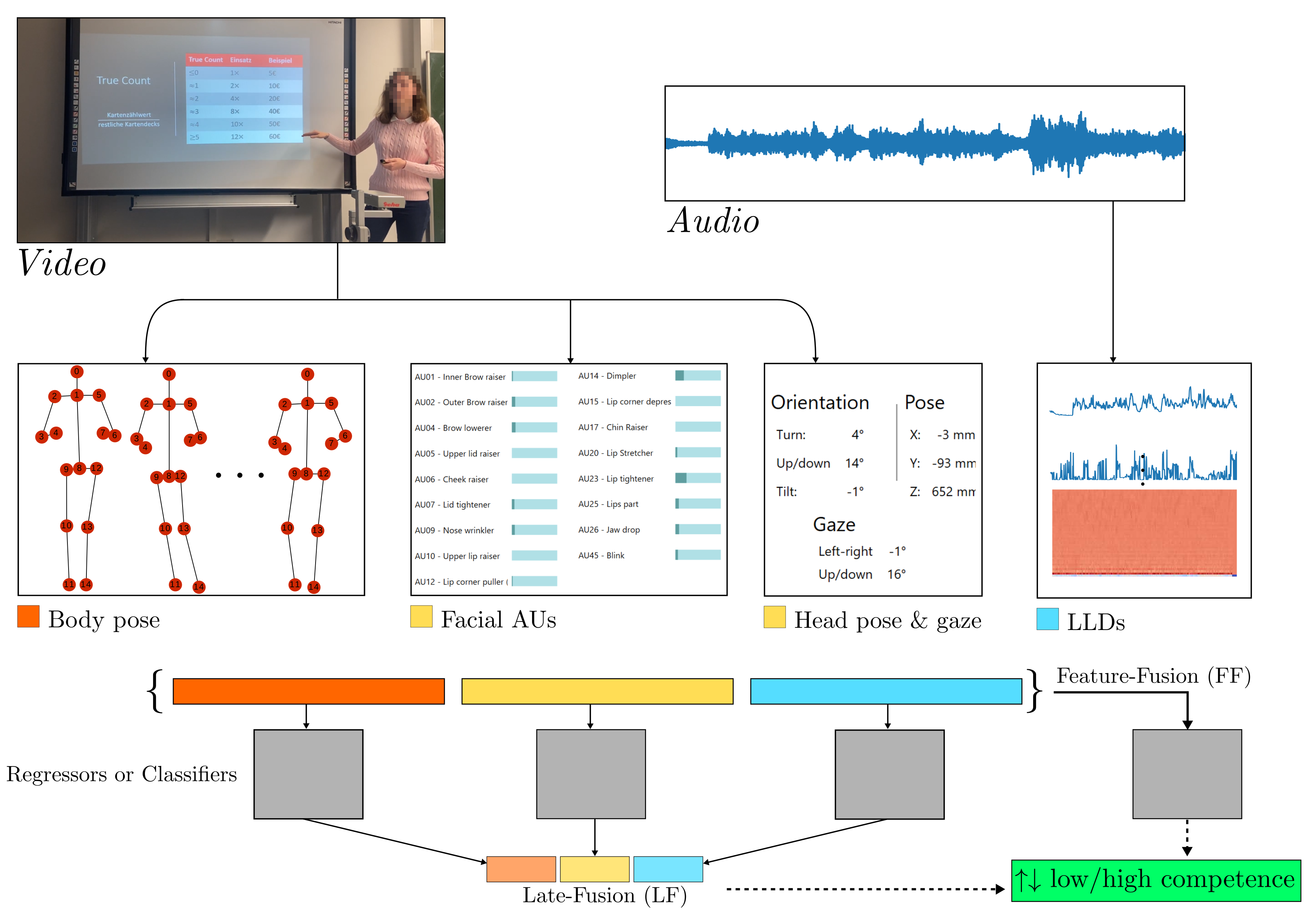}
\end{center}
\caption{Workflow of the proposed method for estimating presentation competence. Our approach uses three main modalities, body pose and facial features from the video and acoustic low-level descriptors (LLDs) from the audio. We investigate different feature fusion (FF) and late fusion (LF) strategies (The used picture is a representative of the dataset but not from the Youth Presents datasets).}
\label{figure:01:workflow}
\end{figure*}

\section{Approach}\label{section:approach}
This section describes our approach to estimating presentation competence from audiovisual recordings of short presentations. We formulated the problem as both classification and regression tasks. When the multimodal aspect of the problem is considered, using different modalities is very crucial. The main features are speech features acquired from acoustic signals and facial and body pose features extracted from visual data.

Figure~\ref{figure:01:workflow} summarizes the main workflow of our method for estimating presentation competence. Using audio and video, we first extract nonverbal features that are relevant for the competence separately. Then, we investigate different fusion strategies, feature-level fusion (FF), and late fusion (LF) using various classifiers and regressors.

\subsection{Nonverbal Feature Extraction}\label{subsection:nonverbal_feature_extraction}

\paragraph{Speech Analysis-based NFs.} Speech analysis is the most popular method to assess presentation performance \cite{Pfister:2011,Chen:2014,Wortwein:2015:a,Cullen:2018}. We used the state-of-the-art acoustic features extraction tool, OpenSMILE \cite{openSmile}, to obtain the extended Geneva Minimalistic Acoustic Parameter Set (eGeMAPS) \cite{Eyben2016}, which constitutes 88 features related to the audio signal.

\paragraph{Facial Analysis-based NFs.} Facial feature extraction consists of the following steps: face detection, facial keypoint estimation, head pose estimation, and FACS action unit occurrence and intensity estimation. We used OpenFace 2.0 \cite{Baltrusaitis:2018} based on Multitask Cascaded Convolutional Networks (MTCNN) \cite{Zhang:2016} for face detection, Convolutional Experts Constrained Local Model (CE-CLM) \cite{Zadeh:2017} for keypoint estimation and perspective n-point (PnP) matching for head pose estimation. AU analysis was performed using Histogram of Oriented Gradients (HOG) and linear kernel Support Vector Machines (SVM) on aligned face patches.

The 43 extracted facial features include the location of the head with respect to the camera in millimetres, rotation angles in radians, eye-gaze directions in radians, the estimated occurrence and intensity of the following action units: Inner brow raiser (AU1), outer brow raiser (AU2), brow lowerer (AU4), upper lid raiser (AU5), cheek raiser (AU6), lid tightener (AU7), nose wrinkler (AU9), upper lid raiser (AU10), lip corner puller (AU12), dimpler (AU14), lip corner depressor (AU15), chin raiser (AU17), lip stretcher (AU20), lip tightener (AU23), lips part (AU25), jaw drop (AU26), and blink (AU45). 

\paragraph{Body Pose NFs.} We examined the use of body pose extracted using the OpenPose algorithm \cite{Cao:2019}. OpenPose estimates the 2-dimensional locations of body joints (i.e., neck, shoulders, arms, wrists, elbows, hips) on video. Skeleton-based data is being used in various problems, for instance, video action recognition, human-computer interaction, and user interfaces, and it also helps to evaluate a presentation. Two items among the TIP labels represent body pose; these are item 10 (effective use of posture) and item 11 (employing gestures convincingly). In the context of presentation competence, using body joints instead of RGB image inputs further eliminates possible subjective bias (i.e., a presenter's visual appearance). We only used 15 joints with locations that were estimated more reliably (depicted in Figure~\ref{figure:01:workflow}).

\begin{figure}
  \centering
   \fontsize{7pt}{7pt}\selectfont
   \def\svgwidth{\columnwidth} 
    \resizebox{0.85\columnwidth}{!}{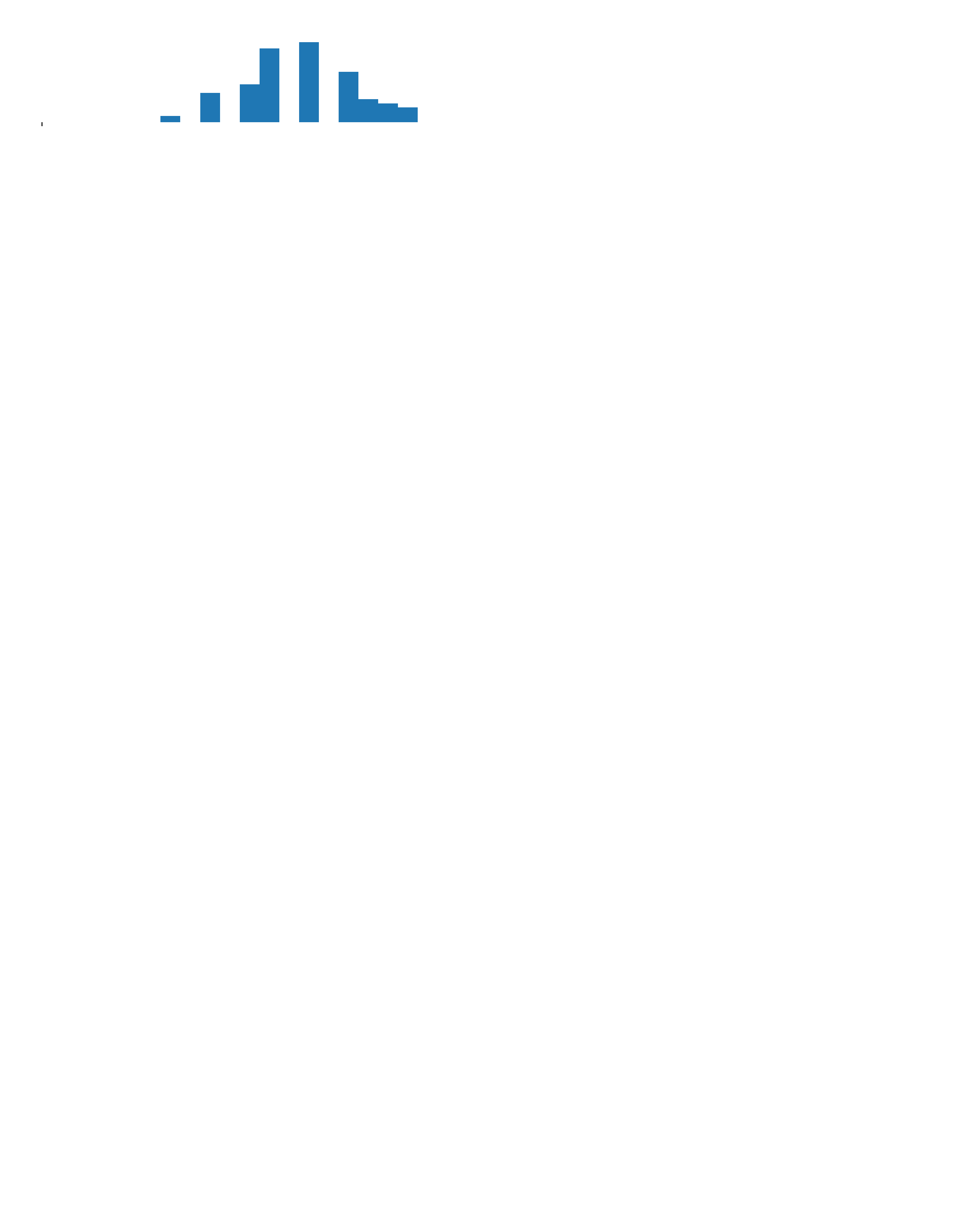}
    \caption{Distribution of body language and voice items in T1 (on the left) and T2 (on the right) data sets. The red line on the overall plots show the median value used from discretization.}
    \label{figure:label_histogram}
\end{figure}

\subsubsection{Global and Local Features} \label{section:GlobalLocal}
Presentation videos are rated using the TIP instrument globally per video, and the average video duration is 3 minutes. However, this duration can contain behavioral cues that contribute to improved presentation competence or vice versa. Understanding these cues in videos is extremely valuable. There are two options to achieve better understanding: use temporally global features or use temporally local features. Global features are extracted from the entire video while local features summarize behaviors during shorter intervals.

A possible use case for presentation analysis is its deployment as a recommender system in the educational domain to help students develop their presentation competence or in the field of therapy to assist people with autism spectrum disorders \cite{Hoque:2013,Tanveer:2015,Tanveer:2016,Tariq:2018}. In this context, localizing parts of a presentation is necessary in order to understand which parts of a presentation are effective in terms of body language and voice competence and which parts are in need of improvement. As continuous annotation of competence in videos is more time-consuming and requires raters with more advanced training, we use local features extracted from 16-second time intervals and use video-level competence items as labels.

Global features directly estimate video level competency. On the other hand, in local features we retrieve the majority vote and the median of predictions in classification and regression, respectively.

\subsection{Classification \& Regression} \label{section:ClassReg}
Presentation competence is a very complicated, multidimensional construct. For instance, among the TIP items shown in \ref{Table:TIP}, addressing an audience, structure, and language require some understanding of a speech's content; this is possible using natural language processing and discourse analysis. In contrast, we focus on items covering body language and voice that can be estimated through nonverbal behavior analysis. 

In this study, we formulated the problem as \textit{i)} a classification or \textit{ii)} a regression task.  While performing regression, we estimated the average of items 10-15 (i.e., the items corresponding to nonverbal communication). In classification, we discretized the ratings of the items 10-15 using the median of their distribution. In that way, we obtained two classes as high or low. In Figure~\ref{figure:label_histogram}, the distribution of items 10-15 is given for T1 and T2 sets of Youth Presents. When T1 and T2 sets were aggregated, the median of presentation competence is 2.83; thus, we used this threshold to discretize continuous values in classification.

In total, four classifiers and regressors: Gradient Boosting (GB) \cite{Friedman:2000}, Decision Tree (DT) \cite{Breiman:1984}, Random Forest (RF) \cite{Breiman:2001}, and Support Vector Machines (SVM) \cite{Cortes:1995,Chang:2011} were applied. These classifiers and regressors were chosen because of their use in the literature for automatic public speaking evaluation (see Section \ref{literature_review} for more details). GB and RF were with 200 estimators. In SVM, rbf kernels and C=10 were used. In all classifiers and regressors, the data is first normalized by removing the mean and scaling to a unit variance of the training set.

\subsection{Data Fusion} \label{section:Fusion}
Estimating presentation competence necessitates understanding several modalities at the same time. Presentation competence items also cover different aspects of nonverbal behaviors. Thus, the fusion of various modalities is highly essential in the performance of presentation estimation. We compared feature-level and late fusion. Feature level fusion combines speech, face, and body pose features and trains a single classifier whereas late fusion combines decision scores of classifiers trained on different feature modalities.

We used two main fusion methods: feature fusion (FF) and late fusion (LF). In feature fusion, all input modalities are concatenated in feature-level into a single feature descriptor, and then a single classifier or regressor is trained. In late fusion, we used the median rule, product rule, and sum rule as follows:
\begin{equation}
    \begin{split}
        P_{med}^{(i)}  & = Median(P^i_m) \\
        P_{prod}^{(i)} & = \prod_{m=1}^{K} P^i_m \\
        P_{sum}^{(i)} & = \sum_{m=1}^{K} P^i_m \\
    \end{split}
\end{equation}
where $P$ is the probability retrieved from each classifier for class $i$. In regression tasks, we applied only median rule on the continuous predicted values from all input modalities.

\section{Experimental Analysis \& Results}\label{section:experimental_analysis}
\label{section:Results}

In classification tasks, the evaluation metrics are accuracy, precision, recall, and the average F1-score. They are given as follows:
\begin{equation}
    \begin{split}
        Accuracy  &= \frac{TP+TN}{TP+TN+FP+FN} \\
        Precision &= \frac{TP}{TP+FP}; \ \  Recall = \frac{TP}{TP+FN} \\
        F1{-}score  &= \frac{2 \cdot Precision\cdot Recall}{Precision+ Recall} \\
    \end{split}
\end{equation}
where TP, TN, FP and FN stand for true positive, true negative, false positive and false negative, respectively. Positive class represents the high presentation competence while negative class represents the low presentation competence.

For the regression task, we used Mean Squared Error (MSE; Eq.~\ref{eq3}) and Pearson Correlation Coefficients ($p-values$ lower than 0.001; Eq.~\ref{eq4}). 
\begin{equation}\label{eq3}
MSE = (\frac{1}{n})\sum_{i=1}^{n}(y_{i} - x_{i})^{2} \\
\end{equation}
\begin{equation}\label{eq4}
\rho = \frac{\text{cov}(X,Y)}{\sigma_x \sigma_y}, 
r = \frac{{}\sum_{i=1}^{n} (x_i - \overline{x})(y_i - \overline{y})}
{\sqrt{\sum_{i=1}^{n} (x_i - \overline{x})^2(y_i - \overline{y})^2}}
\end{equation}
where $\rho$; pearson coefficient value of 1 represents a perfect positive relationship, -1 a perfect negative relationship, and 0 indicates the absence of a relationship between variables $x$ and $y$ (i.e., distributions $X$ and $Y$) while $r$ is the Pearson Correlation estimate. In our case $Y$ and $n$ are the ground-truth and number of samples, respectively.

\begin{table}[ht!] 
\caption{Estimating presentation competence using global and local features as a classification task in T1 data set (N=160). Each result is the average and the standard deviation of 10-fold cross validation. GB, DT, RF, SVM, FF, LF, S, F, BP stand for Gradient Boosting, Decision Tree, Random Forest, Support Vector Machines, feature fusion, late fusion, speech, face and body pose, respectively. The best results are emphasized in bold-face.}
\label{table:cls:baseline}
\begin{minipage}[t]{.48\linewidth}
    \begin{adjustbox}{width=0.98\columnwidth}
    \begin{tabular}{lllllll }
    \toprule
     \multicolumn{6}{l}{\textbf{Classification} (global features)} \\
    \midrule
    Modalities & & Method & Accuracy & Precision & Recall & F1-score  \\
    \midrule
    \textbf{S}peech  &  & GB  & 65.62 & 66.23 & 77.36 & 70.66 \\
                     &  & DT  & 58.13 & 61.75 & 64.49 & 61.61 \\
                     &  & RF  & 66.25 & 66.98 & 76.49 & 70.40 \\
                     &  & SVM & 63.75 & 67.95 & 69.79 & 67.16  \\
    \midrule
    \textbf{F}ace    &   & GB  & 57.50 & 61.03 & 64.44 & 61.51 \\
                     &   & DT  & 63.75 & 68.17 & 66.06 & 65.46 \\
                     &   & RF  & 60.62 & 62.19 & 71.68 & 65.62 \\
                     &   & SVM & 60.00 & 65.42 & 65.07 & 62.48 \\
    \midrule
    \textbf{B}ody Pose  &  & GB  & 63.12 & 65.57 & 72.69 & 66.88 \\
                        &  & DT  & 53.12 & 57.27 & 59.76 & 56.19 \\
                        &  & RF  & 64.38 & 65.91 & 71.26 & 67.87 \\
                        &  & SVM & 59.38 & 60.83 & 68.29 & 63.33 \\
    \midrule
    \emph{Fusion, GB}  & \multicolumn{2}{l}{FF}       & 71.25 & 73.06 & 78.08 & 74.54 \\
    (\textbf{S+F+BP})  & \multicolumn{2}{l}{LF (med)}  & 66.25 & 66.82 & 79.95 & 71.13 \\
                       & \multicolumn{2}{l}{LF (prod)} & 66.88 & 67.68 & 79.19 & 71.78 \\
                       & \multicolumn{2}{l}{LF (sum)}  & 66.25 & 66.67 & 79.43 & 71.39 \\
    \bottomrule
    \end{tabular}
    \end{adjustbox}
\end{minipage}
\begin{minipage}[t]{.48\linewidth}
    \begin{adjustbox}{width=0.98\columnwidth}
    \begin{tabular}{lllllll }
    \toprule
     \multicolumn{6}{l}{\textbf{Classification} (local features, majority voting in video)} \\
    \midrule
    Modalities & & Method & Accuracy & Precision & Recall & F1-score  \\
    \midrule
    \textbf{S}peech  &  & GB  & 65.62 & 66.35 & 74.46 & 69.49 \\
                     &  & DT  & 60.00 & 61.91 & 68.24 & 63.69 \\
                     &  & RF  & 62.50 & 63.85 & 72.28 & 66.84 \\
                     &  & SVM & 60.00 & 63.79 & 68.44 & 63.94 \\
    \midrule
    \textbf{F}ace    &   & GB  & 62.50 & 63.75 & 76.49 & 68.45 \\
                     &   & DT  & 56.88 & 58.94 & 60.49 & 58.86 \\
                     &   & RF  & 63.12 & 63.52 & 77.48 & 68.97 \\
                     &   & SVM & 60.00 & 65.25 & 64.81 & 63.00 \\
    \midrule
    \textbf{B}ody Pose  &  & GB  & 60.00 & 61.98 & 67.81 & 63.34 \\
                        &  & DT  & 58.13 & 64.01 & 58.49 & 58.58 \\
                        &  & RF  & 61.88 & 62.31 & 75.46 & 67.36 \\
                        &  & SVM & 65.62 & 67.94 & 71.22 & 67.77 \\
    \midrule
    \emph{Fusion, GB}  & \multicolumn{2}{l}{FF}         & 65.62 & 68.33 & 75.82 & 69.80 \\
    (\textbf{S+F+BP})  & \multicolumn{2}{l}{LF (med)}   & 66.25 & 65.47 & 84.38 & 72.46 \\
                       & \multicolumn{2}{l}{LF (prod)}  & 66.25 & 66.78 & 83.67 & 72.45 \\
                       & \multicolumn{2}{l}{LF (sum)}   & 65.62 & 65.89 & 83.67 & 72.06 \\
    \bottomrule
    \end{tabular}
    \end{adjustbox}
\end{minipage}
\end{table}

\subsection{The-Same-Dataset Analysis}
\label{section:theSameData}
The results reported in this section include the-same-dataset analysis such that we divided the T1 set into 10-fold so each resulting fold contains a similar number of samples belonging to high or low classes. Meanwhile, if a video (or video segment) belonging to one person exists in a training fold that person is not occurring in the corresponding test fold. Thus, the aforementioned 10-fold cross validation is person-independent. 

Tables Table~\ref{table:cls:baseline} and Table~\ref{table:reg:baseline} report the classification and regression results respectively for each nonverbal feature set (speech, face, and body pose), both individually and when they are fused with feature fusion and late fusion strategies.

Presentation competence labels represent the entire video. However, the ability to estimate presentation competence in shorter time intervals is highly desirable because it can point to areas of low and high competence and would allow researchers to use the proposed methods as part of a self-regulatory tool. We chose 16-second intervals as an alternative to the global features, where all features were aggregated during the entirety of each video. Considering that we work on 3-4 minutes presentations, using 16-second intervals is a good balance and allows having 10-15 sequences from a video on average. 

The classification results in Table~\ref{table:cls:baseline} show that using 16-second intervals does not cause an explicit drop in classification performance. In contrast, it even further improved the accuracy and F1-scores when facial features were used. In most of the feature and classifier combinations, the best performing classifiers are GB and RF.

In feature and late fusion (Table~\ref{table:cls:baseline}), GB classifiers are used as a reference. The performance of FF is 5.63\% better in accuracy than the best performing classifier when speech features were used 65.62\%. The performances of different late fusion approaches are on par. Using multi-modal NFs, i.e., the fusion of all NF sets resulted in an increase in classification performance while the best results were obtained with FF. 

When the effect of using global or local features is examined in terms of the best performance of each NFs group, there is no statistically significant difference. However, there is a clear performance gain when local features were used in some feature/classifier combinations, for instance, +6.24\% in body pose features and SVM classifier and +5\% in facial features and GB classifier. 

The results of regression tasks are depicted in Table~\ref{table:reg:baseline}. In regression, speech features are the best performing one when single modality was used. In contrast to the classification task where using local features improved the performance in some feature and classifier combinations, using local features resulted in correlation between the ground truth labels and predictions dropped significantly. In fusion, feature fusion (FF) and late fusion (LF; by using only median rule) were compared in GB regressors. FF performs better than LF (with Pearson r of 0.61 and 0.56 in both global and local features, respectively), and also beyond the best performing single modalities. 

\begin{table}[ht!] 
\caption{Estimating presentation competence using global and local features as a regression task in T1 data set (N=160). MSE is reported as the mean and the standard deviation of 10-fold cross validation. Pearson correlation coefficients are between the estimated and the ground truth values of all samples. All $p-values$ are lower than $0.001$.
GB, DT, RF, SVM, FF, LF, S, F, BP stand for Gradient Boosting, Decision Tree, Random Forest, Support Vector Machines, feature fusion, late fusion, speech, face and body pose, respectively.}
\label{table:reg:baseline}
\begin{minipage}[t]{.48\linewidth}
    \centering
    \begin{adjustbox}{width=0.9\linewidth}
    \begin{tabular}{llllc }
    \toprule
    \multicolumn{5}{l}{\textbf{Regression}} \\
    \midrule
    Modalities & & Method & MSE  & Pearson \emph{r} \\
    \midrule
    \textbf{S}peech   & & GB  & 0.09 $\pm$ 0.02 & 0.52 \\
                      & & DT  & 0.18 $\pm$ 0.07 & 0.26 \\
                      & & RF  & 0.09 $\pm$ 0.03 & 0.51 \\
                      & & SVM & 0.08 $\pm$ 0.03 & 0.56 \\
    \midrule
    \textbf{F}ace     & & GB  & 0.11 $\pm$ 0.02 & 0.37 \\
                      & & DT  & 0.18 $\pm$ 0.06 & 0.30 \\
                      & & RF  & 0.10 $\pm$ 0.02 & 0.44 \\
                      & & SVM & 0.10 $\pm$ 0.03 & 0.46 \\
    \midrule
    \textbf{B}ody Pose  & & GB  & 0.11 $\pm$ 0.04 & 0.37 \\
                        & & DT  & 0.20 $\pm$ 0.03 & 0.19 \\
                        & & RF  & 0.11 $\pm$ 0.03 & 0.36 \\
                        & & SVM & 0.12 $\pm$ 0.04 & 0.39 \\
    \midrule
    \emph{Fusion, GB}  & \multicolumn{2}{l}{FF}          & 0.08 $\pm$ 0.02 & 0.61 \\
    (\textbf{S+F+BP})  & \multicolumn{2}{l}{LF (med)}    & 0.09 $\pm$ 0.03 & 0.51 \\
    \bottomrule
    \end{tabular}
    \end{adjustbox}
\end{minipage}
\begin{minipage}[t]{.48\linewidth}
    \centering
    \begin{adjustbox}{width=0.9\linewidth}
    \begin{tabular}{llllc }
    \toprule
    \multicolumn{5}{l}{\textbf{Regression} (local-features, averaged per video)} \\
    \midrule
    Modalities & & Method & MSE  & Pearson \emph{r} \\
    \midrule
    \textbf{S}peech   & & GB  & 0.09 $\pm$ 0.04 & 0.50 \\
                      & & DT  & 0.12 $\pm$ 0.05 & 0.35 \\
                      & & RF  & 0.10 $\pm$ 0.04 & 0.43 \\
                      & & SVM & 0.11 $\pm$ 0.04 & 0.38 \\
    \midrule
    \textbf{F}ace    & & GB  & 0.11 $\pm$ 0.04 & 0.31 \\
                     & & DT  & 0.14 $\pm$ 0.07 & 0.19 \\
                     & & RF  & 0.10 $\pm$ 0.03 & 0.40 \\
                     & & SVM & 0.11 $\pm$ 0.04 & 0.32 \\
    \midrule
    \textbf{B}ody Pose   & & GB  & 0.10 $\pm$ 0.04 & 0.43 \\
                         & & DT  & 0.13 $\pm$ 0.04 & 0.25 \\
                         & & RF  & 0.10 $\pm$ 0.04 & 0.41 \\
                         & & SVM & 0.11 $\pm$ 0.05 & 0.32 \\
    \midrule
    \emph{Fusion, GB}  & \multicolumn{2}{l}{FF}          & 0.08 $\pm$ 0.03 & 0.56 \\
    (\textbf{S+F+BP})  & \multicolumn{2}{l}{LF (med)}    & 0.09 $\pm$ 0.03 & 0.54 \\
    \bottomrule
    \end{tabular}
    \end{adjustbox}
\end{minipage}
\end{table}

\subsection{The Cross-Dataset Analysis}
\label{section:theCrossData}
The cross-dataset analysis refers to using a model trained on $T1$ set to predict the $T2$ set (shown as $T1 \rightarrow T2$ ).
The $T1 \rightarrow T2$ setting is important in order to investigate the generalizability of a model trained with the employed NFs. 
Additionally, we also tested the importance of rhetorical settings on the automated analysis, and, in particular, the effect of variations in presentation topics and the speakers' background as related to the presented topic.
We recall here that the presentations in $T1$ set each cover different topics while $T2$ covers presentations on the same topic. In the T1 set, the speakers picked their presentation topic and had more time to prepare (implying that they might build a better background regarding the topic) while in T2 the presentation topic was assigned to the speakers with limited time to prepare.

We applied the same classifier, regressors, global, local features, FF and LF fusions for the cross-dataset experiments as in Section \ref{section:theSameData}. The entire T1 set was used as a training set, and the models were evaluated on 10 folds of T2 data set. Cross data set classification and regression results are given in Table~\ref{table:cls:cross} and Table~\ref{table:reg:cross}.

We should note that the T1 and T2 settings are different in terms of rhetorical setting; however, the T2 data set is the subset of T1 participants. Thus, our cross-dataset evaluation is not person-independent. In classification, global features' performance in all modalities is considerably lower than in the same dataset results. This is a clear sign of the effect of presentation setting on the estimation of competence.

The gap between global and local features is more visible in cross-dataset evaluation. The performance of speech and face deteriorated when local features were used. On the other hand, body pose features exhibited a 10-30\% improvement in accuracy when local features composed of 16-second sequences were used. Even the weakly supervised nature of video-wise labeling is considered and the entire T1 data set is also limited in size (N=160). Using shorter trajectories further increased the size of the training set (N=1.8K) and yielded even better results than person-independent performance on the same data set, particularly in body pose features and fusion.

Looking into the cross-dataset regression results in Table~\ref{table:reg:cross} using GB regressors, the use of local features negatively impacted performance (more than the performance drop from global to local features in Table~\ref{table:reg:baseline}) with the exception of speech features which performed even better than global features. This being the case, when the problem is formulated as regression the use of local features (shorter than the length of actual labels) negatively impacts both the same-dataset and cross-dataset evaluation and should be avoided. In all regression methods, gradient boosting regression with speech features is the best performing method that also retains a high correlation (varying from 0.50 to 0.61) with ground truth labels.

\begin{table}[ht!] 
    \caption{Classification across tasks. All models were trained on the entire T1 set and evaluated on T2 set. The average of accuracy of F1-scores in 10-folds were reported.}
    \centering
    \resizebox{\linewidth}{!}{%
    \begin{tabular}{lllll }
    \toprule
    Modalities/Method  & GB & DT & RF & SVM \\
    \emph{(global features)} & Accuracy / F1-score & Accuracy / F1-score & Accuracy / F1-score & Accuracy / F1-score \\
    \midrule
    \textbf{S}peech  & 57.89 / 56.26 & 56.89 / 51.99  & 57.00 / 48.72 &  66.89 / 53.83 \\
    \textbf{F}ace    & 40.56 / 47.00 & 52.56 / 55.81 & 46.11 / 51.02 &  64.67 / 55.93 \\
    \textbf{B}ody \textbf{P}ose  & 48.11 / 54.98 & 62.67 / 56.40 & 50.33 / 56.70 &  49.33 / 54.21 \\
    \midrule
    (\textbf{S+F+BP})  &         &   &   &   \\
    FF          & 48.33 / 47.55  & 42.67 / 42.58 & 49.22 / 51.81 &  57.00 / 51.52 \\
    LF (med)    & 49.22 / 57.08  & 62.44 / 61.21 & 49.22 / 53.88 &  63.56 / 58.91 \\
    LF (prod)   & 52.44 / 57.48  & 69.22 / 45.31 & 49.22 / 55.80 &  59.33 / 54.07 \\
    LF (sum)    & 50.33 / 56.75  & 62.44 / 61.21 & 49.22 / 55.80 &  59.33 / 54.07 \\
    \midrule
    Modalities/Method & GB & DT & RF & SVM \\
    (local features) & Accuracy / F1-score & Accuracy / F1-score & Accuracy / F1-score & Accuracy / F1-score \\
    \midrule
    \textbf{S}peech             & 44.89 / 59.02 & 55.78 / 68.24  & 51.44 / 64.30 &  59.22 / 72.95 \\
    \textbf{F}ace               & 60.33 / 75.09 & 68.00 / 79.76  & 68.00 / 80.67 &  70.22 / 82.39 \\
    \textbf{B}ody \textbf{P}ose & 79.22 / 88.31 & 52.78 / 66.59  & 79.22 / 88.31 &  78.11 / 87.56 \\
    \midrule
    (\textbf{S+F+BP})           &                &               &               &                \\
    FF                          & 64.89 / 78.54  & 57.11 / 72.29 & 66.89 / 77.74 &  74.78 / 85.49 \\
    LF (med)                    & 71.44 / 83.24  & 56.00 / 70.99 & 72.56 / 83.99 &  77.00 / 86.90 \\
    LF (prod)                   & 68.11 / 80.75  & 40.44 / 50.15 & 70.44 / 82.41 &  78.11 / 87.65 \\
    LF (sum)                    & 68.11 / 80.75  & 56.00 / 70.99 & 70.44 / 82.41 &  78.11 / 87.65 \\
    \bottomrule
    \end{tabular}
    }
    \label{table:cls:cross}
\end{table}

\begin{table}[ht!] 
    \caption{Gradient Boosting (GB) regression across task. All models were trained on the entire T1 set and evaluated on T2 set. MSE is reported as the average and standard deviation of 10-folds. Pearson correlation coefficients are between the estimated and the ground truth values of all samples in T2 data set (N=91). All $p-values$ are lower than $0.05$.}
    \centering
    \begin{tabular}{lcc}
    \toprule
    Modalities &  MSE  & Pearson \emph{r} \\
    \midrule
    \multicolumn{3}{l}{\textbf{Global features} } \\
    Speech         & 0.12 $\pm$ 0.04 & 0.45 \\
    Face           & 0.14 $\pm$ 0.04 & 0.25 \\ 
    Body Pose      & 0.19 $\pm$ 0.04 & 0.21 \\
    FF             & 0.13 $\pm$ 0.04 & 0.41 \\
    LF (med)       & 0.13 $\pm$ 0.03 & 0.43 \\
    \midrule
    \multicolumn{3}{l}{\textbf{Local features} } \\
    Speech         & 0.12 $\pm$ 0.01 & 0.51 \\
    Face           & 0.18 $\pm$ 0.04 & 0.01 \\ 
    Body Pose      & 0.18 $\pm$ 0.01 & 0.08 \\
    FF             & 0.16 $\pm$ 0.01 & 0.25 \\
    LF (med)       & 0.14 $\pm$ 0.01 & 0.43 \\
    \bottomrule
    \end{tabular}
    \label{table:reg:cross}
\end{table}

\subsection{Which feature is better?}\label{section:whichFea}
When all three modalities, speech, face, and body pose features, were compared, speech features outperformed face and body pose features in the same dataset evaluation. With the exception of the DT classifier or regressor, speech features consistently performed better than the other two features in both classification and regression tasks. The fact that decision trees are weaker learning models than GB, RF, and SVM is one possible explanation. Overall, speech features appear to be the most dominant nonverbal cues to estimate presentation competence.

When visual nonverbal features, face and body pose, were considered, body pose features were more efficient in most cases. The use of local features further improved the performance (for instance, GB, RF, and SVM in cross dataset classification, DT and SVM in same-dataset classification). These results indicate that finer granularity of body postures leads to a better understanding of competence. Beyond that, the labeling of prototypical body postures can further improve classification and regression performance. 

\section{Conclusion}\label{section:conclusion}
This study presented an analysis of computer vision and machine learning methods to estimate presentation competence. We used audiovisual recordings of a real-world setting, the Youth Presents Presentation Competence Datasets. The dataset contained different challenges: presentation time and free selection of topics in the T1 data set and limited preparation time and predetermined topics and preparation materials in the T2 data set. We used a recently proposed instrument, Tübingen Instrument for Presentation Competence (TIP), and validated that it could be used to train automated models to estimate presentation competence.

We formulated presentation competence estimation as classification and regression tasks and conducted nonverbal analysis of presenters' behaviors. The modalities used were speech (affective acoustic parameters of voice), facial features (head pose, gaze direction, and facial action units), and body pose (the estimated locations of body joints). Classification and regression methods were gradient boosting (GB), decision trees (DT), random forests (RF), and support vector machines (SVM).

In the-same-dataset, evaluation (T1), our classification approach reached 71.25\% accuracy and 74.54\% F1-score when early fusion was applied. In regression, we could reach a mean squared error of 0.08 and Pearson correlation of 0.61. In both settings, the feature-level fusion strategy performed better than late fusion, combining the scores of separate models.

Training and testing in different rhetorical settings still seems difficult. Even though the T2 set contains different speeches from the same persons, having enough time to prepare and the ability to freely select a presentation topic impacts classification and regression performance.

Estimating presentation competence in a finer granularity is a key priority in the development of recommender systems that sense the nonverbal behaviors and give feedback to the presenter. The use of shorter sequences (16-seconds) and subsequent statistics of nonverbal features aggregated in these shorter time windows does not deteriorate performance, but, rather, helps significantly in cross-dataset evaluation.

\paragraph*{Limitation.} Automated methods to estimate presentation competence can be an essential asset in education. Considering the importance of effective and successful presentation competence in academic and professional life, such systems can help students more effectively gain those competencies and provide additional support for teachers. However, the use of automated methods must comply with ethical standards and should only be deployed with the users' consent.

From the perspective of fairness, in contrast to the raw image input in many computer vision tasks, we used processed nonverbal behavioral features. For instance, the datasets and algorithms that estimate attentional features (head pose and gaze direction), emotional features (facial expressions and action units), and body pose contain various subjects representative of different demographics. Still, dataset and algorithmic fairness are highly critical issues in the current data-driven learning approaches. Beyond nonverbal feature extraction tasks, a more diverse and large-scale dataset is necessary to accurately model all behavioral differences (i.e., cultural variations) while delivering a presentation. 

\paragraph*{Future Work.}
In future work, we plan to increase the data scale to model all behavioral variances more accurately. The personalization of presentation competence models and development of recommender systems and user interfaces are also among future research topics.

\paragraph*{\textbf{Acknowledgements.}}
Ömer Sümer is a member of LEAD Graduate School \& Research Network, which is funded by the Ministry of Science, Research and the Arts of the state of Baden-Württemberg within the framework of the sustainability funding for the projects of the Excellence Initiative II. This work is also supported by Leibniz-WissenschaftsCampus Tübingen ``Cognitive Interfaces''. Cigdem Beyan is supported by the EU Horizon
2020 Research and Innovation Programme under project AI4Media (GA No. 951911).

\bibliographystyle{plain}
\bibliography{main} 

\end{document}